\documentclass[letterpaper]{article} 
\usepackage{aaai2027}  
\usepackage[hyphens]{url}  
\usepackage{graphicx} 
\usepackage{natbib}  
\usepackage{caption} 
\usepackage{algorithm}
\usepackage{algorithmic}

\usepackage{newfloat}
\usepackage{listings}
\DeclareCaptionStyle{ruled}{labelfont=normalfont,labelsep=colon,strut=off} 
\floatstyle{ruled}
\newfloat{listing}{tb}{lst}{}
\floatname{listing}{Listing}

\usepackage{booktabs}

\usepackage{multirow}
\title{One Perception, All Maneuvers: Directional Traffic Signal Understanding for Maneuver-Level Signal Intent Prediction
}
\author{
    Written by AAAI Press Staff\textsuperscript{\rm 1}\thanks{With help from the AAAI Publications Committee.}\\
    AAAI Style Contributions by Peter Patel Schneider,
    Sunil Issar,\\
    J. Scott Penberthy,
    George Ferguson,
    Hans Guesgen,
    Francisco Cruz\equalcontrib\corresponding,
    Marc Pujol-Gonzalez\equalcontrib\corresponding
}
\affiliations{
    \textsuperscript{\rm 1}Association for the Advancement of Artificial Intelligence\\

    1101 Pennsylvania Ave, NW Suite 300\\
    Washington, DC 20004 USA\\
    proceedings-questions@aaai.org
}

\title{One Perception, All Maneuvers: Directional Traffic Signal Understanding for Maneuver-Level Signal Intent Prediction}
\author {
    Ang Zou\textsuperscript{\rm 1}\equalcontrib,
    Runzhe Zheng\textsuperscript{\rm 1},
    Zhigang Li\textsuperscript{\rm 2},
    Zhen Yang\textsuperscript{\rm 2},
    Xia Han\textsuperscript{\rm 2},
    Xuewei Li\textsuperscript{\rm 3}\corresponding,
    Zequn Qin\textsuperscript{\rm 1}\corresponding,
    Xi Li\textsuperscript{\rm 1}\corresponding
}
\affiliations {
    \textsuperscript{\rm 1}College of Computer Science and Technology, Zhejiang University\\
    \textsuperscript{\rm 2}BYD Company Limited\\
    \textsuperscript{\rm 3}School of Electronic and Information Engineering, Shanghai DianJi University\\
    \{12621098, 22621160, qinzequn, xilizju\}@zju.edu.cn, \{li.zhigang9, xia.han3 \}@byd.com, yangzhensean@gmail.com, xueweili@sdju.edu.cn
}

\begin{document}

\maketitle

\begin{abstract}
Traffic lights are a key regulatory signal for autonomous driving at urban intersections, yet existing traffic signal perception is still predominantly formulated as instance-level detection or color recognition. Such formulations identify where traffic lights are and what colors they display, but leave a critical semantic gap before downstream planning: which ego maneuver is controlled by each visible signal and what dynamic permission the signal expresses for that maneuver. 

In this paper, we formulate \textbf{Directional Traffic Signal Understanding}, a decision-oriented task that predicts structured signal states for straight, left-turn, right-turn, and U-turn maneuvers from a front-view image. Each state contains the associated signal color and signal-implied passability. Based on OpenLane-V2, we provide a direction-level benchmark with maneuver-level supervision and metrics for color recognition, passability, full-frame consistency, and safety-critical errors. A direction-aware baseline combines global context, localized traffic-light evidence, and maneuver-specific representations. Experiments show that direction-level modeling improves passability prediction over image-level classifiers and detection-oriented pipelines, particularly at complex multi-signal intersections. The resulting representation provides a direct and interpretable traffic-signal interface for downstream planning together with topology, route, and surrounding-agent information.
\end{abstract}


\section{Introduction}

\begin{figure*}[t]
\centering
\includegraphics[width=1.0\textwidth]{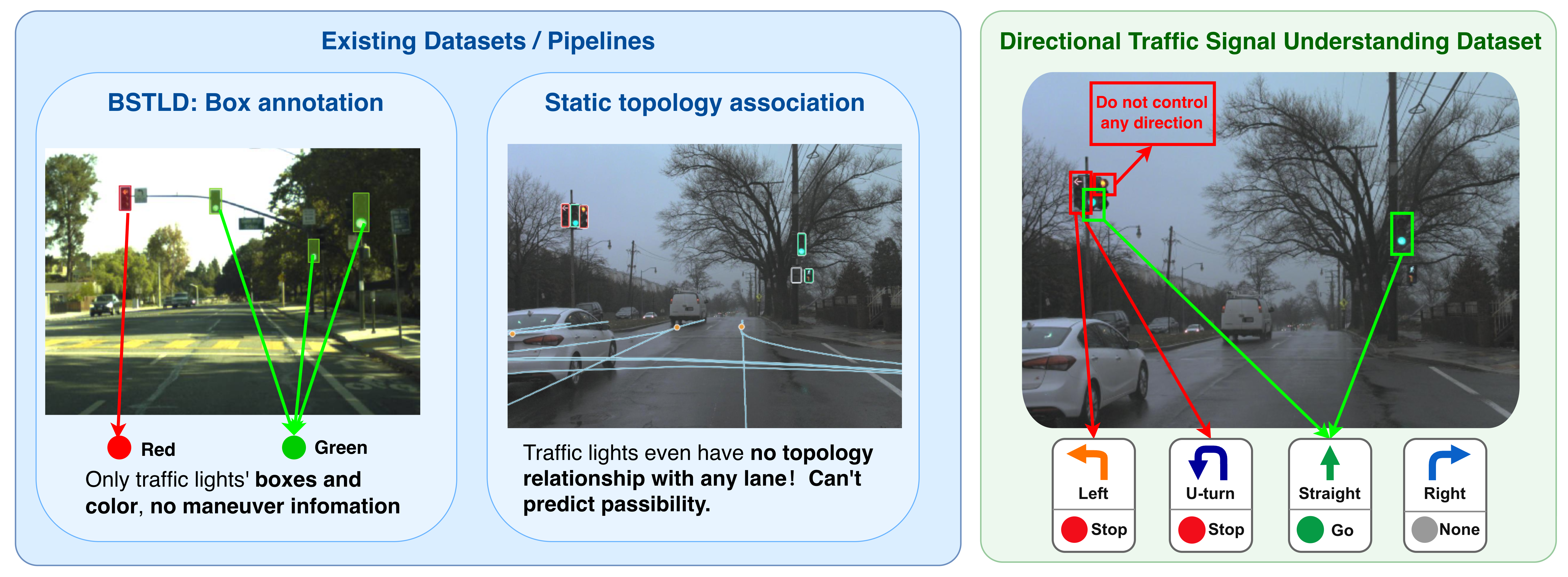} 
\caption{Overview of the proposed Directional Traffic Signal Understanding task.
Conventional traffic light recognition focuses on detecting signal instances and recognizing their colors, which leaves an ambiguity between visible signals and ego-vehicle maneuvers.
We reformulate traffic signal perception as maneuver-level passability understanding, where a model predicts structured signal states for going straight, turning left, turning right, and making a U-turn. The \texttt{none} label indicates that no dedicated signal
controls the corresponding direction, whose signal-implied
passability is treated as \texttt{go}. }
\label{fig2}
\end{figure*}

Urban intersections are among the most rule-intensive and safety-critical scenarios for autonomous driving~\citep{grigorescu2020survey,janai2020computer,badue2021self}. When an ego vehicle approaches a signalized intersection, static lane topology and route connectivity describe which maneuvers are geometrically feasible, while dynamic traffic signals convey the current control state of these maneuvers~\citep{wang2023openlane,li2024lanesegnet,fu2024topologic}. Reliable intersection driving therefore requires not only perceiving visible traffic lights, but also understanding their implications for the ego vehicle's possible movements.

Existing traffic-light perception generally involves traffic-light localization, visual-state recognition, arrow recognition, signal-to-lane association, and traffic-rule interpretation~\citep{jensen2016vision,behrendt2017deep,fregin2018driveu,wang2021traffic,jang2017traffic,hirabayashi2019traffic}. These tasks are commonly studied or implemented as separate stages, whose outputs are subsequently combined with topology information or handcrafted rules to obtain maneuver-level signal states.

This fragmented formulation leaves a gap between traffic-light perception and downstream planning. Detecting a traffic light and recognizing its color do not directly indicate which ego maneuver it controls. At complex intersections, multiple signals may appear simultaneously and correspond to different lanes or movements. The system must therefore integrate signal detection, state recognition, directional interpretation, and signal-to-maneuver association before producing the structured signal information required by planning.

We formulate these capabilities around a common objective: \textbf{maneuver-level traffic-signal intent understanding}. Here, the predicted state represents the permission implied specifically by traffic signals, rather than the final feasibility of executing a maneuver. A downstream planner can further combine this signal-level intent with lane topology, route constraints, surrounding agents, and other traffic rules.

To this end, as shown in Figure~\ref{fig2}, we propose \textbf{Directional Traffic Signal Understanding}. Given a front-view image, the task predicts structured traffic-signal states for four ego-centric maneuvers: going straight, turning left, turning right, and making a U-turn. Each state contains the associated signal color and its corresponding signal-implied permission. The main challenge is to identify relevant traffic-light evidence and associate it with the correct maneuver, especially in scenes with multiple signal groups, directional arrows, small or occluded lights, and signals belonging to adjacent lanes or distant intersections~\citep{jensen2016vision,fregin2018driveu}.

We establish a direction-level benchmark based on OpenLane-V2~\citep{wang2023openlane}. During offline construction, traffic-element annotations and topology-aware scene information are used to establish signal-to-maneuver supervision~\citep{li2026graph}. At inference time, models directly predict maneuver-level color and signal-implied permission states from front-view images without requiring precise static topology as input. We further provide a unified evaluation protocol and a direction-aware baseline for this task. This formulation reduces reliance on multi-stage association pipelines and provides a compact, interpretable, and planning-oriented traffic-signal representation that can be directly integrated with other scene and route information. 

Our contributions are summarized as follows:

\begin{itemize}
    \item To the best of our knowledge, we are the first to formulate
    \textbf{Directional Traffic Signal Understanding} as a unified task
    that integrates traffic-light localization, state recognition,
    directional interpretation, and signal-to-maneuver association for
    maneuver-level passability prediction.

    \item We establish a \textbf{direction-level intersection benchmark} based on OpenLane-V2,  together with a unified evaluation protocol covering signal recognition, maneuver-level passability, structured consistency, and safety-critical errors.

    \item We develop a \textbf{direction-aware baseline} that combines global scene context, localized traffic-light evidence, and maneuver-specific representations, and conduct extensive experiments to evaluate the proposed task.
\end{itemize}
\begin{figure*}[!t]
    \centering
    \includegraphics[
        width=1.0\textwidth,
        height=0.8\textheight,
        keepaspectratio
    ]{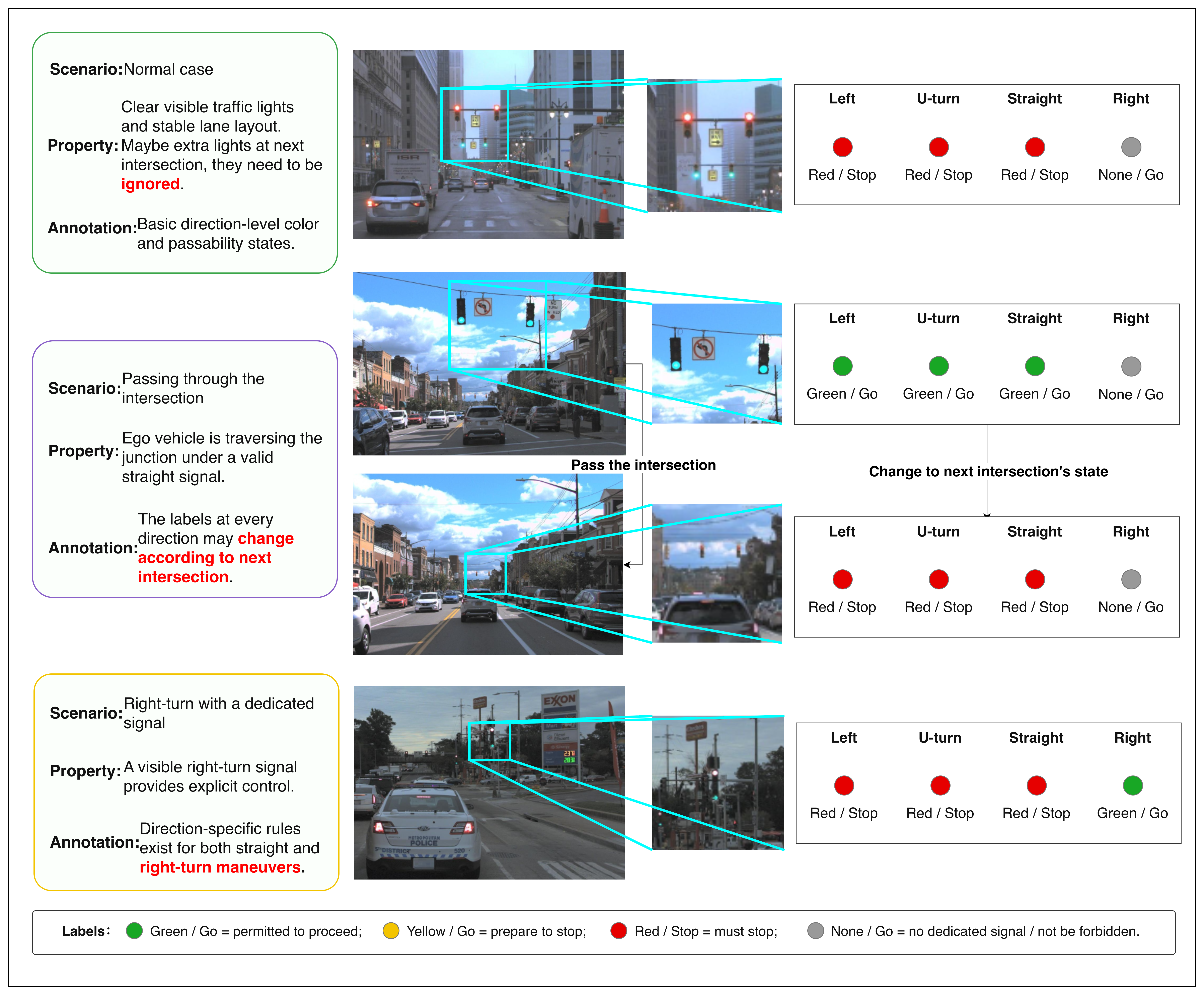}
    \caption{Representative scenarios and direction-level annotations in the proposed passability benchmark. The examples illustrate a standard signalized intersection, label transitions when the ego vehicle passes through an intersection, and a dedicated right-turn signal. For each frame, the benchmark provides color and passability states for left-turn, U-turn, straight, and right-turn maneuvers, while excluding traffic lights that control adjacent lanes or distant intersections.}
    \label{fig:dataset}
\end{figure*}
\section{Related Work}

\subsection{Traffic-Light Perception}

Traffic-light perception traditionally includes localization,
state recognition, directional-arrow recognition, and relevance
estimation. Early methods rely on hand-crafted color and shape
features~\citep{de2009real,gong2010recognition}, while modern
traffic-light systems build on generic two-stage and one-stage
detectors~\citep{ren2015faster,redmon2016you} and multi-scale
feature representations~\citep{lin2017feature} to improve the
recognition of small and distant signals~\citep{behrendt2017deep,
fregin2018driveu,wang2021traffic}.
Map-assisted methods further associate detected signals with
the ego lane or planned route~\citep{jang2017traffic,
hirabayashi2019traffic}. More recently, TLS-Assist introduces
a modular traffic-light and traffic-sign recognition layer that
extracts safety-critical regulatory information for downstream
driving agents~\citep{schmidt2025enhancing}.

Despite this progress, most methods produce instance-level
detections or an overall relevance estimate. They do not
directly provide the color and traffic-signal permission for
each candidate ego maneuver, leaving an additional association
and rule-conversion stage before planning.

\subsection{Traffic Topology and Signal--Lane Association}
Large-scale autonomous-driving datasets support perception,
mapping, tracking, and forecasting~\citep{geiger2012we,
caesar2020nuscenes,sun2020scalability,yu2020bdd100k,
wilson2023argoverse}, but do not organize traffic-signal
annotations around maneuver-level permissions. OpenLane-V2
introduces lane--lane and lane--traffic-element topology
relationships~\citep{wang2023openlane}, followed by methods
that improve lane representation and topology reasoning through
lane segments, graph modeling, and interpretable
pipelines~\citep{li2024lanesegnet,fu2024topologic,
li2026graph}. Traffic regulations have also been incorporated
into vectorized HD maps~\citep{chang2025driving}.

More recent methods improve topology construction through
endpoint detection or one-stage joint prediction of lanes,
traffic elements, and their relations~\citep{fu2026topopoint,
li2025reusing}. Although these methods can derive
maneuver-level signal states from predicted topology, they
still require intermediate lane and relation reconstruction.
Our task instead directly evaluates the traffic-signal state
associated with each ego maneuver.

\subsection{Decision-oriented Driving Understanding}

Recent autonomous-driving research increasingly organizes
perception around downstream decision requirements, including
visual question answering, rule-aware reasoning, topology
understanding, occupancy-based planning, safety-oriented
accident prediction, and driving simulation~\citep{
qian2024nuscenes,lu2025can,wu2025language,
yang2025driving,wang2024deepaccident,dauner2024navsim,
cai2026driving,ma2026drivecombo}.

Our work differs from prior studies by providing a unified formulation
of traffic-light perception. We regard these traffic-light perception tasks as complementary capabilities toward
a common objective: understanding the signal-implied passability of ego
maneuvers. This formulation shifts traffic-light perception from a set
of fragmented object-level tasks to a unified, decision-oriented
understanding problem.

\begin{figure*}
    \centering
    \includegraphics[width=1\linewidth]{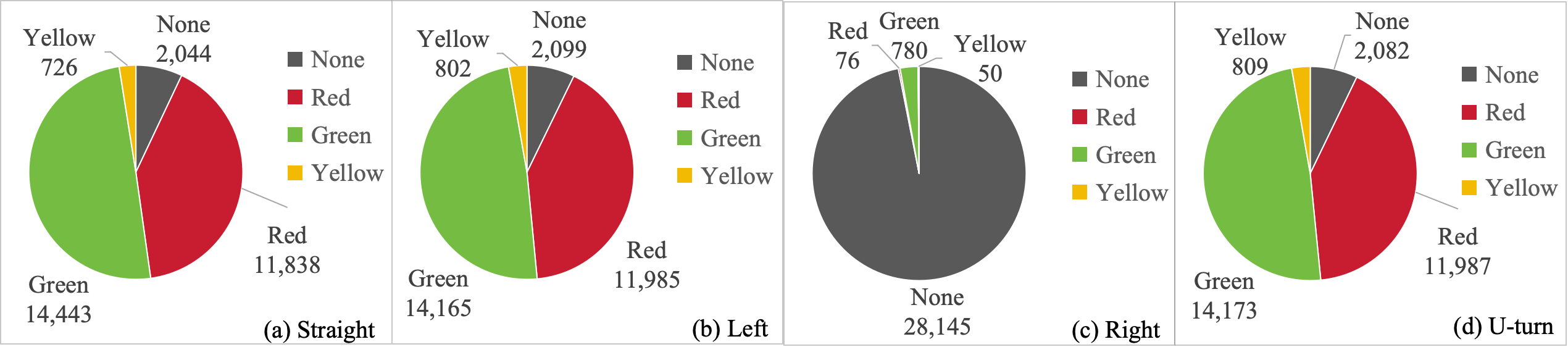}
    \caption{Color-state data distributions for straight, left-turn, right-turn, and U-turn maneuvers.}
    \label{fig:dist}
\end{figure*}
\section{Maneuver-level Passability Understanding}
Previous discussion reveals a missing interface between traffic light perception and downstream planning. Existing methods mainly recognize visible signal instances, while maneuver-level permissions are usually inferred later through static topology, map priors, or handcrafted association rules. To bridge this gap, we introduce \textbf{Directional Traffic Signal Understanding}, a decision-oriented perception task that directly predicts dynamic traffic-signal semantics for ego-vehicle maneuvers.
\subsection{Task Formulation}
Let $I_i$ denote the $i$-th front-view image captured by the ego vehicle at an intersection. The objective is to predict a structured directional traffic signal state:
\begin{equation}
Y_i =
\left(
y_i^s,
y_i^l,
y_i^r,
y_i^u
\right).
\end{equation}
where $s$, $l$, $r$, and $u$ correspond to straight, left-turn, right-turn, and U-turn maneuvers, respectively. Each directional state is represented as:
\begin{equation}
    y_i^{d} = (c_i^{d}, p_i^{d}), \quad d \in \mathcal{D}.
\end{equation}

where $\mathcal{D}=\{s,l,r,u\}$. Here, $c_i^{d}$ denotes the traffic signal color associated with direction $d$, and $p_i^{d}$ denotes the passability state of that direction. A model $f_\theta$ takes $I_i$ as input and predicts:
\begin{equation}
\hat{Y}_i
=
f_\theta(I_i)
=
\left(
\hat{y}_i^s,
\hat{y}_i^l,
\hat{y}_i^r,
\hat{y}_i^u
\right).
\end{equation}

This formulation differs from instance-level traffic light recognition in two aspects. First, the prediction target is ego-centric: outputs are organized by driving maneuvers rather than detected objects. Second, the supervision is structured: the model must reason about all directions within the same intersection scene. As a result, the task evaluates not only visual recognition, but also the model's ability to associate traffic signal evidence with maneuver-level passability.

\subsection{Evaluation Protocol}
\label{sec:evaluate}
A benchmark for Directional Traffic Signal Understanding should measure both perceptual correctness and decision relevance. We therefore design a model-agnostic evaluation protocol with four complementary metrics: color accuracy, passability accuracy, exact match, and a unified Traffic Signal Score. We further report safety-critical errors to characterize failure modes that are particularly important for autonomous driving.

\paragraph{Color and Passability Accuracy: }
We use color accuracy $\mathrm{Acc}_{\mathrm{color}}$ and passability accuracy $\mathrm{Acc}_{\mathrm{pass}}$ to evaluate direction-level prediction from two complementary perspectives. Color accuracy measures whether the model correctly predicts the traffic signal color associated with each ego maneuver, including the \texttt{none} category. It reflects the model's ability to recognize direction-level visual signal states.

Passability accuracy measures whether the model correctly predicts the decision-level state of each maneuver direction, namely whether the ego vehicle should go or stop for a specific maneuver. Compared with color accuracy, passability accuracy is more directly related to downstream planning, since it evaluates whether visual traffic signal evidence is correctly converted into maneuver-level driving semantics.

\paragraph{Exact Match: }Direction-wise metrics evaluate each maneuver independently, but they do not fully capture whether a model understands the entire intersection state. We therefore define Exact Match $\mathrm{EM}$, which requires all evaluated directional states in a frame to be correct:
\begin{equation}
\mathrm{EM} = 
\frac{1}{N}
\sum_{i=1}^{N}
\mathcal{I}
\left(
\forall d\in\mathcal{D},
\hat{c}_i^{d}=c_i^{d}
\land
\hat{p}_i^{d}=p_i^{d}
\right).
\end{equation}

where $\mathcal{I}(\cdot)$ is the indicator function. 

$\mathrm{EM}$ is a strict structured metric. A frame is considered correct only when the model predicts all
direction-level color and passability states correctly. It therefore reflects full-frame consistency rather than isolated directional accuracy.

\paragraph{Traffic Signal Score:}To summarize benchmark-level performance, we define the Traffic Signal Score $\mathrm{TSS}$ as:
\begin{equation}
\mathrm{TSS}=
\frac{1}{3}
\left(
\mathrm{Acc}_{\mathrm{color}}
+
\mathrm{Acc}_{\mathrm{pass}}
+
\mathrm{EM}
\right).   
\end{equation}

$\mathrm{TSS}$ jointly measures visual signal recognition, maneuver-level passability prediction, and structured scene-level consistency. We use it as the primary aggregate metric, while retaining individual metrics for detailed diagnosis.

\paragraph{Safety-critical Error Metrics: }Autonomous driving applications require distinguishing different types of errors. In particular, predicting a non-passable direction as passable is more safety-critical than predicting a passable direction as non-passable. We therefore report false-go and false-stop errors.

The overall false-go rate is defined as:
\begin{equation}
\mathrm{False}\texttt{-}\mathrm{Go}=
\frac{
\sum_i\sum_{d\in\mathcal D}
\mathcal{I}(p_i^{d}=\texttt{stop}
\land
\hat{p}_i^{d}=\texttt{go})
}{
\sum_i \sum_{d\in\mathcal D}
\mathcal{I}(p_i^{d}=\texttt{stop})
}.
\end{equation}

The overall false-stop rate is defined as:

\begin{equation}
    \mathrm{False}\texttt{-}\mathrm{Stop}=
\frac{
\sum_i \sum_{d\in\mathcal D}
\mathcal{I}(p_i^{d}=\texttt{go}
\land
\hat{p}_i^{d}=\texttt{stop})
}{
\sum_i \sum_{d\in\mathcal D}
\mathcal{I}(p_i^{d}=\texttt{go})
}.
\end{equation}

False-go errors reflect unsafe permissive predictions, while false-stop errors reflect overly conservative behavior. Reporting both provides a more complete view of model reliability under decision-oriented traffic signal understanding.

\section{Data Construction}

\subsection{Benchmark Overview}

To support Directional Traffic Signal Understanding, we construct a
direction-level passability benchmark based on OpenLane-V2. Representative samples are shown in Figure~\ref{fig:dataset}. Unlike
conventional traffic-light datasets organized around individual signal
instances, our benchmark uses ego-vehicle maneuvers as the annotation
unit. It unifies traffic-light localization, state recognition,
directional interpretation, and signal-to-maneuver association through
a common prediction target: the current passability of each ego
maneuver.

For each front-view frame, we annotate four directions,
$\mathcal{D}=\{\texttt{straight},\texttt{left},
\texttt{right},\texttt{U-turn}\}$. Each direction is represented by
$y^d=(c^d,p^d),$$d\in\mathcal{D},$
where $c^d\in\{\texttt{none},\texttt{red},
\texttt{green},\texttt{yellow}\}$ denotes the associated signal state,
and $p^d$ denotes whether the corresponding maneuver is permitted.

The benchmark contains 23,904 training frames and 5,147 validation
frames, corresponding to 95,616 and 20,588 direction-level states,
respectively. As shown in Figure~\ref{fig:dist}, straight, left-turn,
and U-turn directions contain dense valid signal supervision, whereas
right-turn states are more frequently labeled as \texttt{none}, since
many right-turn maneuvers are not controlled by a dedicated visible
signal.

\subsection{Maneuver-level Annotation Construction}

OpenLane-V2 provides front-view images together with traffic elements,
lane structures, and topology annotations. We retain frames near
signalized intersections that contain sufficient visual and structural
evidence for determining maneuver-level traffic states, while removing
ordinary road segments and severely ambiguous samples.

For each retained frame, traffic-light annotations, including color,
arrow type, spatial location, and traffic-element semantics, are
associated with ego-centric maneuvers using lane layout and topology
information. Signals controlling adjacent lanes, opposite traffic, or
distant intersections are excluded. The associated signal state is then
converted into a structured color and passability label for each
maneuver.

If no reliable signal can be associated with a direction, its color
state is labeled as \texttt{none}. Through this conversion, the original
object-level annotations are reorganized into a unified maneuver-level
output space directly aligned with downstream driving decisions.

\subsection{Quality Control}

We combine automatic consistency checks with manual verification.
Automatic checks identify missing labels, invalid color--passability
combinations, and conflicts with the original traffic-element
annotations. Manual verification focuses on challenging cases,
including multi-signal intersections, directional arrow lights, small
or occluded signals, and ambiguous signal-to-maneuver associations.

Labels are corrected when the relevant signal can be reliably
identified; otherwise, the corresponding direction is marked as
\texttt{none}. This procedure reduces annotation noise while preserving
the complexity of real-world intersection scenes.

\section{Baseline Model}
\begin{figure}[!t]
    \centering
    \includegraphics[
        width=\columnwidth,
        height=0.45\textheight,
        keepaspectratio
    ]{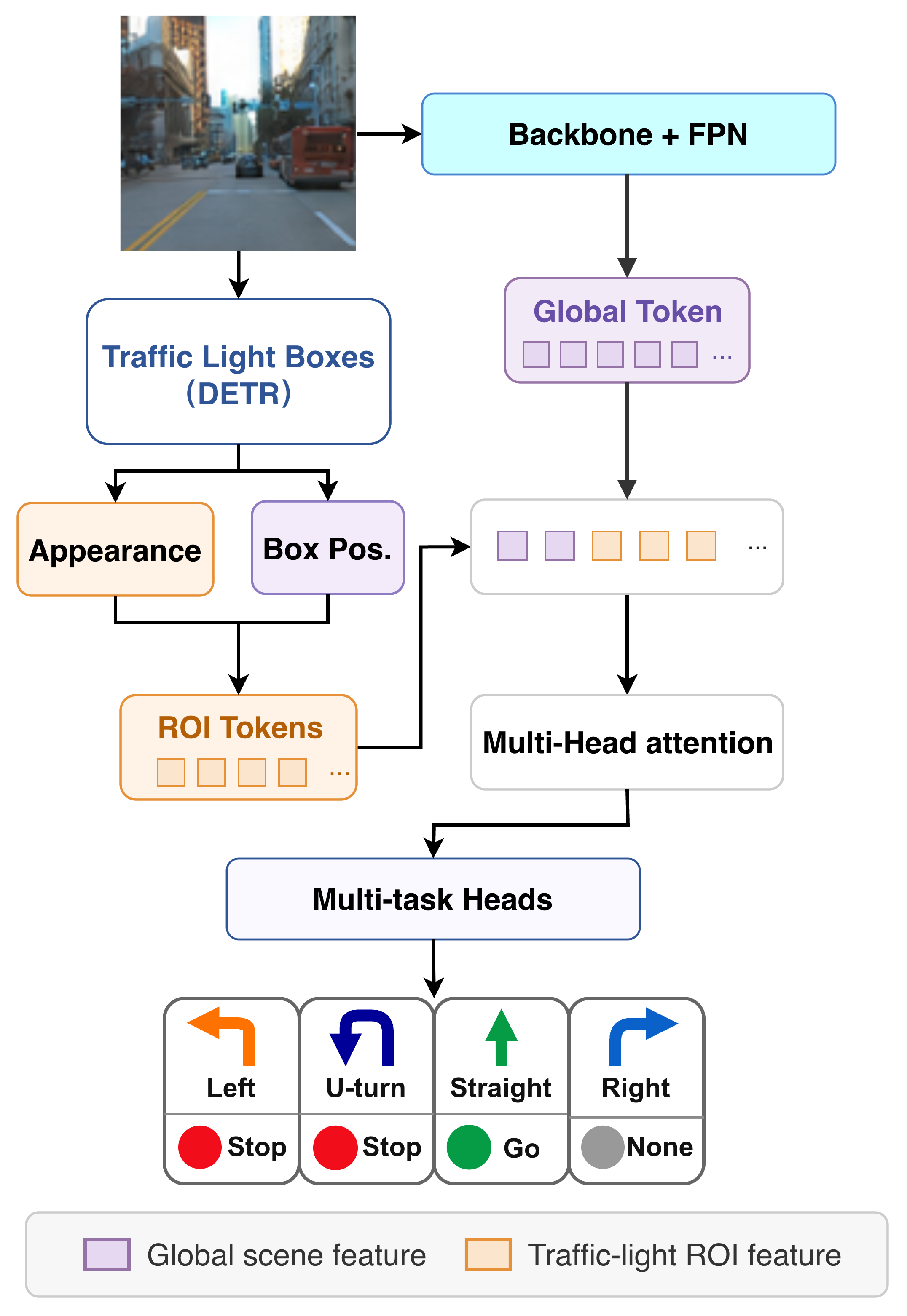}
    \caption{Overview of the proposed framework. Global scene features and
    localized traffic-light ROI features are combined to predict color and
    passability for four ego maneuvers: left, U-turn, straight, and right.}
    \label{fig:model}
\end{figure}

To provide a reproducible reference for the proposed benchmark, we design a simple direction-aware baseline, as shown in Figure~\ref{fig:model}. Rather than building a full planning system, the baseline focuses on the central capability of Directional Traffic Signal Understanding: predicting maneuver-level color and passability states directly from front-view visual input. This design allows us to examine whether dynamic traffic rules can be learned without relying on precise static road topology at inference time.

\subsection{Overall Architecture}

Given a front-view image $I$, a visual
backbone with a Feature Pyramid
Network (FPN) extracts multi-scale visual
features:
\begin{equation}
    \mathcal{F}=\mathrm{FPN}(\mathrm{Backbone}(I)).
\end{equation}

A global token $g$ is obtained by pooling the feature pyramid
to represent the overall intersection context. Meanwhile, a
lightweight DETR-style detector generates traffic-light
bounding boxes. For each detected box $\hat b_j$, we extract
its localized appearance and spatial information to construct
an ROI token:
\begin{equation}
t_j =
\phi_r\!\left(\mathrm{RoIAlign}(\mathcal{F},\hat b_j)\right)
+\phi_b(\hat b_j).
\end{equation}
where function $\phi$ is a linear projection network. The global token and all traffic-light ROI tokens are combined
into a unified token sequence:
\begin{equation}
T=[g,t_1,\ldots,t_K].
\end{equation}

\subsection{Direction-aware Prediction}

To associate the visible traffic-light evidence with individual
ego maneuvers, we introduce four learnable direction queries,
one for each of straight, left-turn, right-turn, and U-turn.
Each query attends to the unified token sequence:
\begin{equation}
h^d=\mathrm{MHA}(q^d,T,T),
\qquad d\in\mathcal{D}.
\end{equation}
The resulting direction-specific representation $h^d$ is then
fed into two prediction heads:
\begin{equation}
\hat c^d=\mathrm{Head}_{\mathrm{color}}(h^d),
\qquad
\hat p^d=\mathrm{Head}_{\mathrm{pass}}(h^d).
\end{equation}
The color head predicts
$\{\texttt{none},\texttt{red},\texttt{yellow},\texttt{green}\}$,
while the passability head predicts
$\{\texttt{stop},\texttt{go}\}$.

This design combines global scene context with localized
traffic-light evidence and directly produces structured states
for all four maneuvers, without requiring handcrafted
signal-to-maneuver assignment at inference time.

\begin{table}[t]
\centering
\scriptsize
\setlength{\tabcolsep}{2.5pt}
\begin{tabular}{lcccccc}
\toprule
Method
& Acc$_{\mathrm{color}}$ $\uparrow$
& Acc$_{\mathrm{pass}}$ $\uparrow$
& EM $\uparrow$
& TSS $\uparrow$
& False-Go $\downarrow$
& False-Stop $\downarrow$ \\
\midrule

ResNet-18
& 0.8514 & 0.8977 & 0.7253 & 0.8248 & 0.3224 & 0.0335 \\

ResNet-50
& 0.8775 & 0.9142 & 0.7745 & 0.8554 & 0.2651 & \textbf{0.0297} \\

ResNet-101
& 0.8795 & 0.9165 & 0.7702 & 0.8554 & 0.2209 & 0.0405 \\

ViT-B/16
& 0.8698 & 0.9054 & 0.7702 & 0.8485 & 0.1612 & 0.0738 \\

Swin-Tiny
& 0.8916 & 0.9266 & 0.8019 & 0.8734 & 0.1831 & 0.0391 \\

DeiT-Small
& 0.8634 & 0.9062 & 0.7528 & 0.8408 & 0.2387 & 0.0485 \\

\midrule

Ours
& \textbf{0.9019}
& \textbf{0.9356}
& \textbf{0.8031}
& \textbf{0.8802}
& \textbf{0.1372}
& 0.0416 \\

\bottomrule
\end{tabular}
\caption{
Comparison with representative generic visual models on
Subset-A. All models use the same four-maneuver output space
and a unified input resolution. Higher is
better for the first four metrics, while lower is better for
False-Go and False-Stop. The best result in each column is
highlighted in \textbf{bold}.
}
\label{tab:main_safety_results}
\end{table}

\begin{table*}[t]
\centering
\small
\setlength{\tabcolsep}{3.2pt}
\begin{tabular}{lccrrrrrr}
\toprule
Method
& Params
& Setting
& Acc$_{\mathrm{color}}$ $\uparrow$
& Acc$_{\mathrm{pass}}$ $\uparrow$
& EM $\uparrow$
& TSS $\uparrow$
& False-Go $\downarrow$
& False-Stop $\downarrow$ \\
\midrule

Qwen2.5-VL-7B-Instruct
& 7B
& SFT-LoRA
& \underline{0.8779} & \underline{0.9126} & \underline{0.7892} & \underline{0.8599} & 0.2814 & \textbf{0.0268} \\

MiniCPM-V 4.5
& 8B
& SFT-LoRA
& 0.8712 & 0.9043 & 0.7888 & 0.8548 & 0.2607 & 0.0442 \\

Llama-3.2-11B-Vision
& 11B
& SFT-LoRA
& 0.8731 & 0.9031 & 0.7830 & 0.8531 & 0.2615 & 0.0454 \\

\midrule

Qwen3.6-35B-A3B
& 35B-A3B
& Zero-shot
& 0.5910 & 0.7780 & 0.1014 & 0.4901 & 0.2745 & 0.2055 \\

Gemma 4 31B
& 31B
& Zero-shot
& 0.4645 & 0.6700 & 0.0104 & 0.3817 & 0.3601 & 0.3206 \\

Nemotron 3 Nano Omni
& 30B-A3B
& Zero-shot
& 0.5437 & 0.6253 & 0.0000 & 0.3897 & \textbf{0.0130} & 0.4877 \\

\midrule

Ours
& \textbf{50.1M}
& Supervised
& \textbf{0.9019}
& \textbf{0.9356}
& \textbf{0.8031}
& \textbf{0.8802}
& \underline{0.1372}
& \underline{0.0416} \\

\bottomrule
\end{tabular}
\caption{
Comparison with recent multimodal large language models on Subset-A.
SFT models are fine-tuned on the training split, while zero-shot models
use no task-specific supervision. The best and second-best results are
shown in \textbf{bold} and \underline{underlined}, respectively.
}
\label{tab:mllm_comparison_subset_a}
\end{table*}

\begin{table*}[t]
\centering
\small
\setlength{\tabcolsep}{4pt}
\begin{tabular}{llcccccc}
\toprule
Method
& Inference mechanism
& Acc$_{\mathrm{color}}$ $\uparrow$
& Acc$_{\mathrm{pass}}$ $\uparrow$
& EM $\uparrow$
& TSS $\uparrow$
& False-Go $\downarrow$
& False-Stop $\downarrow$ \\
\midrule

TLD-READY
& Pictogram mapping
& 0.8563
& 0.9187
& 0.7516
& 0.8422
& 0.3017
& 0.0124 \\

TopoPoint
& Predicted topology
& 0.7180
& 0.8681
& 0.5439
& 0.7100
& 0.5229
& \textbf{0.0096} \\

Reusing Attention / One-Stage
& Reused-attention topology
& 0.6647
& 0.8532
& 0.4762
& 0.6647
& 0.5721
& 0.0138 \\

Ours
& Direct maneuver prediction
& \textbf{0.9019}
& \textbf{0.9356}
& \textbf{0.8031}
& \textbf{0.8802}
& \textbf{0.1372}
& 0.0416 \\

\bottomrule
\end{tabular}
\caption{
Comparison with relevance- and topology-based traffic-light
understanding methods on Subset-A. All methods are evaluated
under the same directional output protocol using the same six
task metrics. TLD-READY derives directional states from
traffic-light and arrow-direction relevance cues, while TopoPoint and One-Stage
use predicted lane--traffic topology. All structured baselines
use approach-signal fallback when explicit directional evidence
is unavailable. The best results are highlighted in \textbf{bold}.
}
\label{tab:recent_methods_subset_a}
\end{table*}

\subsection{Training Objective}

The model is trained with direction-level classification losses
and an auxiliary traffic-light detection loss.

For color prediction, we apply cross-entropy over the four
maneuvers:
\begin{equation}
\mathcal{L}_{\mathrm{color}}
=
\frac{1}{|\mathcal{D}|}
\sum_{d\in\mathcal{D}}
\mathrm{CE}(\hat{c}^d,c^d).
\end{equation}

Similarly, the passability loss is defined as:
\begin{equation}
\mathcal{L}_{\mathrm{pass}}
=
\frac{1}{|\mathcal{D}|}
\sum_{d\in\mathcal{D}}
\mathrm{CE}(\hat{p}^d,p^d).
\end{equation}

The traffic-light detector is trained using the standard DETR
bipartite matching and detection objective, including
classification, box regression, and GIoU losses~\citep{carion2020end}.
We denote the resulting detection loss as
$\mathcal{L}_{\mathrm{det}}$.

The overall training objective is:
\begin{equation}
\mathcal{L}
=
\lambda_c\mathcal{L}_{\mathrm{color}}
+
\lambda_p\mathcal{L}_{\mathrm{pass}}
+
\lambda_d\mathcal{L}_{\mathrm{det}}.
\end{equation}

The direction-level losses optimize the final maneuver states,
while the auxiliary detection loss provides localized
traffic-light evidence for subsequent direction-aware
prediction.



\section{Experiments}

\subsection{Experimental Setup}

We implement all models in PyTorch~\citep{paszke2019pytorch}.
We evaluate ResNet-18/50/101~\citep{he2016deep}, ViT-B/16~\citep{dosovitskiy2020image}, Swin-Tiny~\citep{liu2021swin}, and
DeiT-Small~\citep{touvron2021training} using publicly available pretrained weights. For
backbone ablations, all remaining modules, including the
256-dimensional FPN, detection, fusion, and prediction heads,
are unchanged. Models are trained with
AdamW~\citep{loshchilov2017decoupled} and a cosine learning-rate
schedule.

All methods use identical data splits, preprocessing, and
training settings. Images are resized to a fixed resolution
while preserving their aspect ratio. We report the six metrics
defined in the evaluation protocol.

Multimodal models are evaluated using either SFT-LoRA or
zero-shot prompting with the same output format and parser.
Structured baselines retain their original pipelines, with
outputs converted to our four-maneuver protocol. 

Due to space limitations, the main paper reports results on
Subset-A, while the complete experiments on Subset-B, including
backbone, multimodal model, and structured-method comparisons,
are provided in the \textbf{supplementary material.}

\subsection{Effect of Visual Backbones}

Table~\ref{tab:main_safety_results} compares representative
visual backbones under the same four-maneuver output protocol.
Among the generic models, Swin-Tiny performs best, achieving
0.8916 color accuracy, 0.9266 passability accuracy, 0.8019 EM,
and 0.8734 TSS.

Using the same backbone, our direction-aware model further
improves TSS to 0.8802 and reduces False-Go from 0.1831 to
0.1372, a relative reduction of 25.1\%. ResNet-50 obtains the
lowest False-Stop rate. These results show that backbone
capacity alone is insufficient, while localized signal encoding
and maneuver-aware fusion provide consistent gains in accuracy
and safety-critical reliability.

\subsection{Comparison with Multimodal Large Language Models}

Table~\ref{tab:mllm_comparison_subset_a} compares our compact
task-specific model with both fine-tuned and zero-shot
multimodal large language models. With only 50.1M parameters,
our model uses approximately 140--220$\times$ fewer parameters
than the 7B--11B SFT models, while achieving higher color
accuracy, passability accuracy, EM, and TSS. It also obtains
the lowest False-Go rate among the supervised models.
Although Qwen2.5-VL-7B-Instruct achieves a lower False-Stop
rate, its substantially higher False-Go rate indicates a more
permissive error bias. Conversely, although Nemotron 3 Nano
Omni achieves the lowest False-Go rate, its False-Stop rate
reaches 0.4877 and its EM is zero, indicating an extremely
conservative prediction bias.

The advantage is more pronounced when compared with the 30B--35B
zero-shot models. Despite using approximately 599--699$\times$
fewer parameters, our model substantially outperforms their
zero-shot performance across all four main metrics. These
results suggest that model scale alone is insufficient for
fine-grained signal-to-maneuver reasoning, while a compact
architecture with task-specific inductive biases provides a
more accurate and balanced solution.

\subsection{Comparison with Association- and
Topology-based Methods}
We compare our model with three structured traffic-light
understanding methods. TLD-READY~\citep{polley2024tld} maps detected signal
pictograms to maneuvers with an approach-level fallback.
TopoPoint~\citep{fu2026topopoint} and One-Stage~\citep{li2025reusing} derive maneuver states from predicted
lane--signal topology. These pipelines decompose the task into
detection, state recognition, relevance estimation, topology
association, and rule conversion.

Our formulation unifies these objectives into maneuver-level
traffic intent prediction. It directly predicts the signal color
and passability of each ego maneuver. This design provides
planning-oriented outputs and reduces error propagation between
separate perception and reasoning stages.

As shown in Table~\ref{tab:recent_methods_subset_a}, our model
achieves the highest TSS and EM. It outperforms TLD-READY,
the strongest baseline, by 3.80 points in TSS and 5.15 points in EM, while reducing
False-Go by 54.5\%. These results show
that unified maneuver-level prediction produces more consistent
and safety-aware traffic-intent estimates. 

\subsection{Discussion}
Overall, the results show that maneuver-level traffic-signal understanding benefits from explicitly combining localized signal evidence with direction-aware representations. Generic visual models and large multimodal models capture useful scene context, but are less effective at associating signals with specific ego maneuvers, while topology-based pipelines remain vulnerable to errors in intermediate predictions. Direct maneuver-level prediction therefore provides a more reliable and planning-oriented representation of traffic-signal states.

\section{Conclusion}

We present \textbf{Directional Traffic Signal Understanding}, a benchmark that reorganizes instance-level traffic-light annotations into ego-centric supervision for straight, left-turn, right-turn, and U-turn maneuvers. A direction-aware baseline and unified metrics evaluate signal recognition, structured consistency, and safety-critical errors. Experiments show that maneuver-level modeling outperforms generic image-level approaches and reduces unsafe false-go predictions. Overall, this work provides a \textit{direct traffic-signal interface between perception and downstream planning.}

\bibliography{1}

\end{document}